\documentclass[11pt]{article}
\usepackage[utf8]{inputenc}
\usepackage{fullpage}
\usepackage{authblk}
\usepackage{url}
\usepackage{graphicx}
\usepackage{amsmath}
\usepackage{algorithm} 
\usepackage{algpseudocode} 
\graphicspath{ {./images/} }

\title{Adaptive Stress Testing for Adversarial Learning in a Financial Environment}
\author[1]{Khalid El-Awady\footnote{This work was completed as part of the requirements for Stanford University's AA203 class in the Spring of 2021. The author wishes to thank the course staff for their helpful comments.}}

\date{\today}

\begin{document}
\maketitle
\begin{abstract}
    We demonstrate the use of Adaptive Stress Testing to detect and address potential vulnerabilities in a financial environment. We develop a simplified model for credit card fraud detection that utilizes a linear regression classifier based on historical payment transaction data coupled with business rules. We then apply the reinforcement learning model known as Adaptive Stress Testing to train an agent, that can be thought of as a potential fraudster, to find the most likely path to system failure -- successfully defrauding the system. We show the connection between this most likely failure path and the limits of the classifier and discuss how the fraud detection system's business rules can be further augmented to mitigate these failure modes. 
\end{abstract}

\section{Introduction}
Transactional volume facilitated by branded and domestic payment cards (credit, debit, and prepaid) amounted to \$42.274 trillion in 2019, up 4.2\% over 2018 \cite{Nilson}. Gross fraud losses on these cards reached \$28.65 billion, up 2.9\% from \$27.85 billion in 2018. This amounts to 6.5 cents on every \$100 of payment volume (or 6.5 bps). Further, LexisNexis reports that the payment card industry collectively incurs an additional \$2 for every \$1 of fraud in detection, prevention, and remediation efforts \cite{LexisNexis}. The secular trend in the growth of share of online payments, where fraud rates average 3-4$\times$ those of in-person transactions, is further accelerating these costs.  

Approaches to detecting credit card fraud run the gamut of traditional machine learning techniques, including some of the earliest commercial implementations of neural networks \cite{Samaneh}. Yet the field remains active in both research and industry. This can be attributed to the adaptability of fraudsters who have the luxury of testing and learning with an abundance of cheap stolen user data \cite{ITRC} and in an online environment that is ideal for programmatic and anonymous trials. This requires the financial system to constantly update their defenses and employ novel techniques.

Adversarial learning, in which the actions of the fraudsters are explicitly modeled is a relatively new approach to payment card fraud detection. This approach incorporates an adversary’s potential strategies when attempting to improve defenses. In this report we present an adversarial fraud detection and prevention approach using reinforcement learning. We formulate a model for fraud classification from publicly available credit card fraud data and model the fraudster's actions as a Markov Decision Process (MDP). To devise the fraudster's optimal policy we will explore the application of Adaptive Stress Testing to this environment to see what it might offer.

\section{Related Work}
Most work in the area of adversarial learning in payment card fraud credits Liu and Chawla \cite{LiuChawla} as one of the earliest inspirations for this approach. In that work, the authors present a game theoretic approach to email spam detection. Spam detection shares important similarities with payment card fraud including large class imbalances, a similar amount of high-volume observations that can be generated at low cost, and the presence of malicious actors that want to fool the classifier. This work and its predecessors showed that a spam classifier that takes the adversary’s best actions into account greatly increases their model’s predictive power. 

In \cite{Beling1} and \cite{Beling2} Beling's teams present a reinforcement learning approach to credit card fraud detection that includes an MDP model of fraudster decisions and an active defense by the card issuer. The authors utilize the learned adversary policy to disrupt the adversary’s learning process, altering classifier probability classification thresholds at regular intervals to effectively limit the amount of successful fraud. One drawback of their work is the artificial action and reward spaces they constructed for the fraudster. In \cite{Beling1} the entire data set is split into 3 bulk clusters using Gaussian Mixture Models and the fraudster is allowed to choose as its action a transaction from the cluster that is most profitable. In \cite{Beling2} the action space is limited to a binary choice of a low \$-amount transaction of a high \$-amount transaction. There is also no accounting for the element of time in fraud detection, which is a critical feature of any payment card fraud detection system (e.g., through the imposition of rate limits on transactions).

Recently, Lee et al. \cite{Lee} have a presented a new approach termed Adaptive Stress Testing (AST) that offers a framework for finding the most likely path to a failure event in a system modeled by an MDP. Discovering failures that are most likely to occur is an intriguing enhancement and offers a natural priority for adding incremental mitigation measures. While there is literature describing the application of AST in areas such as collision avoidance systems for air traffic control \cite{Moss} and autonomous driving \cite{Koren}, we have not found any significant literature on its use in a financial setting. 

\section{Problem Statement}
Adversarial learning in fraud prevention has been shown to increase effectiveness over static models that do not account for changing fraudster behavior. Further,  reinforcement learning has been demonstrated to be an effective way to learn a fraudster's attack strategy with an eye towards improving defenses. We show how the performance of a fraud detection system can be enhanced using Adaptive Stress Testing to model a fraudster's behavior and mitigating likely strategies they might employ. 

\section{Approach}
\begin{figure*}[ht]
    \centering
    \includegraphics[scale=0.5]{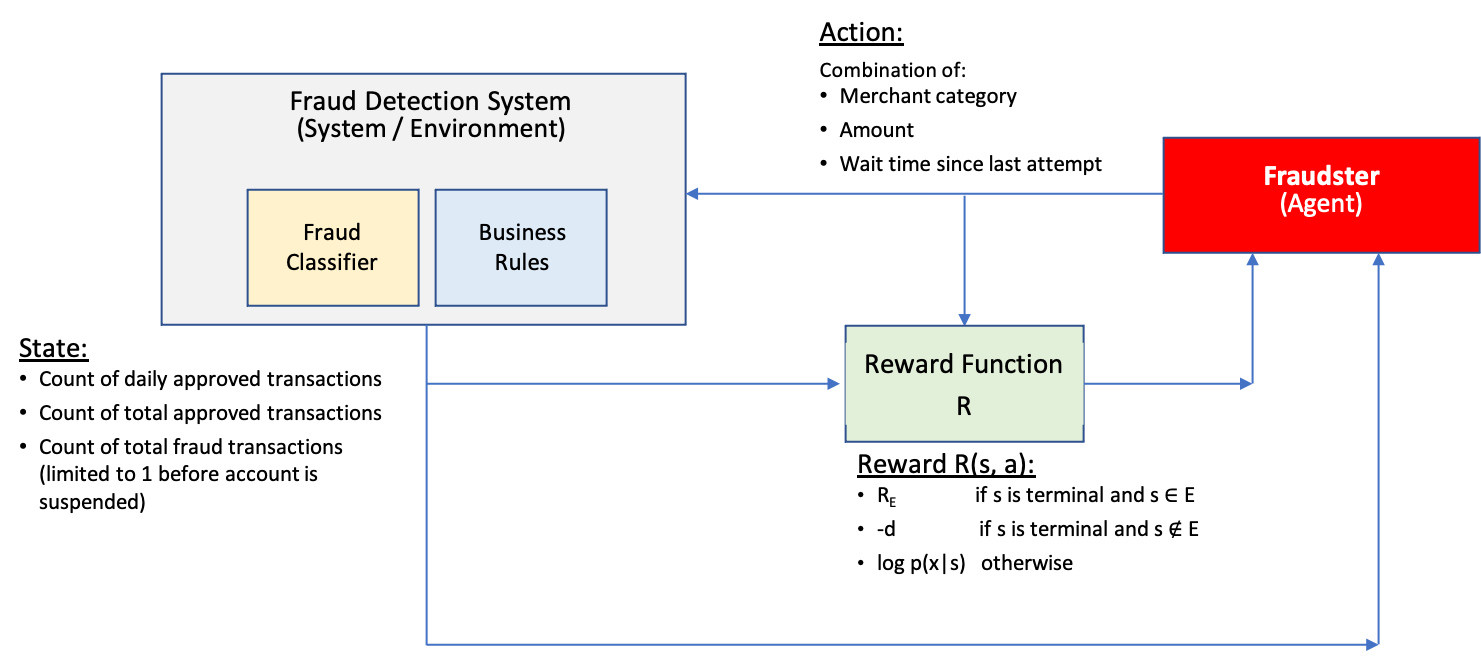}
    \caption{Reinforcement learning model for adversarial learning of the payment card fraud detection system.}
    \label{ASTSystem}
\end{figure*}

Figure \ref{ASTSystem} shows the reinforcement learning model of our problem. The system represents an approver of an attempted payment card transaction (typically  a card issuer). The system receives a request to authorize a payment from a merchant and classifies it as either fraud or not fraud based on learned historical patterns. The system also applies certain business rules such as limits on the number of accepted daily transactions. 

In the AST approach, the learning problem is setup to determine the path of most likely failure (fraud). The agent (fraudster) interacts with the system to learn the 'best' way to commit fraud. In fact this mimics practice closely where it is common for fraudsters to purchase batches of stolen cards test and learn where vulnerabilities are (often with low dollar transaction), and then commit high dollar fraud. 

 \subsection{The System: A Model for Payment Fraud Detection} \label{systemSection}
Modern payment card fraud detection systems utilize at their core a statistical classifier that is trained on historical transaction data, some of which have been reported as fraud. This will use the combined attributes of the available payment data elements to determine whether to approve or decline a payment authorization request. In addition, these systems will also incorporate business rules such as imposing hard limits on amounts of individual transactions and frequency of transacting (amongst other rules). The systems also lean toward conservatism -- at the first strong indication of potential fraud the account is generally suspended and prevented from further transacting. 

In our simplified system model we define the {\bf state} to include 3 variables:
\begin{enumerate}
    \item Cumulative approved transaction count on the card: this tracks the usage of the card and correlates with the age of the card.
    \item Cumulative  daily transaction count on the card: this is a key fraud control parameter and is generally limited to a certain number (say 10) as a stop gap against an undetected fraud run.
    \item Indicator if fraud has been detected: this represents the output of the classifier. It is assumed that the first detected instance of fraud results in immediate suspension of the card. 
\end{enumerate}

To train a fraud classifier we located a appropriate public dataset on Kaggle with sufficient richness and transparency for our exercise: https://www.kaggle.com/kartik2112/fraud-detection. We reduced this to a set of key payment transaction data fields which are either necessary for modeling purposes or are known (from experience) to have a strong correlation to fraud and are subject to fraudster influence: the payment account number, a merchant category index (one of 14 possible values -- see table \ref{AmtStatsTable}), the transaction amount, and the elapsed time interval since last transaction. 

Finally we augment our system with a set of simple {\bf business rules} that reflect common practice:
\begin{enumerate}
    \item The card is suspended at the first detection of a fraudulent transaction.
    \item The card is suspended if the number of daily transactions exceeds 10 transactions.
\end{enumerate}

\begin{figure*}[ht]
    \centering
    \includegraphics[scale=0.45]{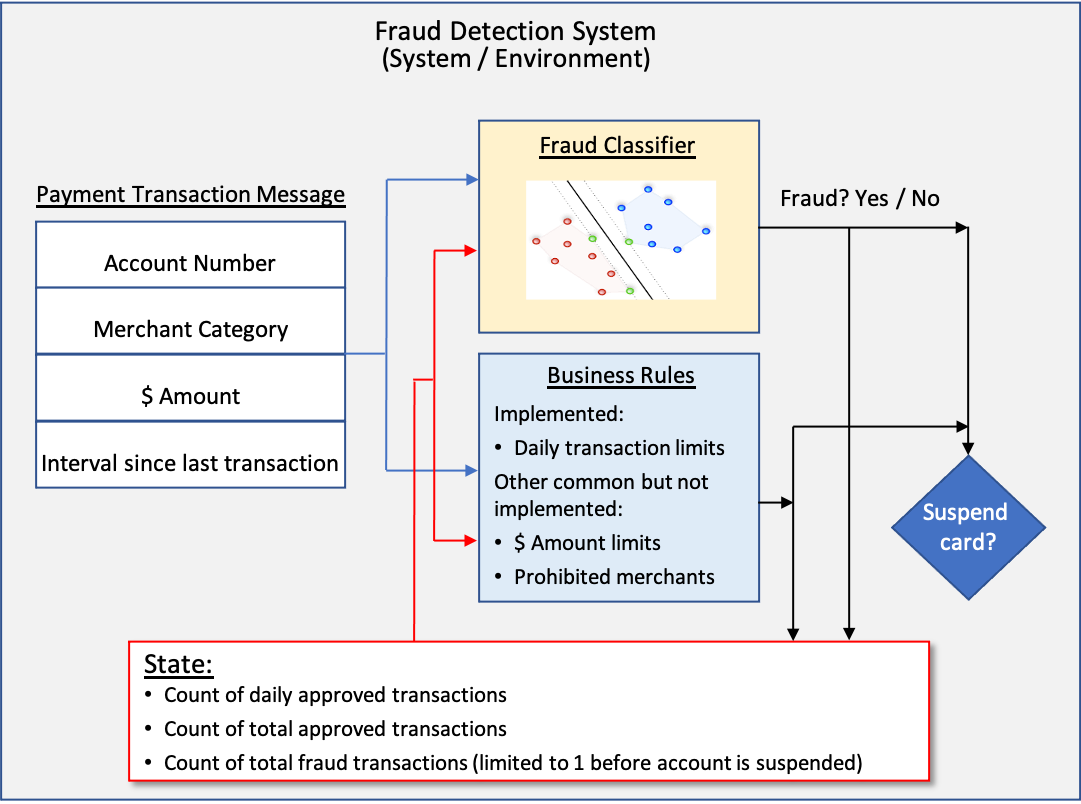}
    \caption{The fraud detection system.}
    \label{System}
\end{figure*}

The system is depicted graphically in Figure \ref{System}. Note that with these assumptions a fraud episode (using a single payment card) ends whenever the system detects fraud or if the fraudster performs 10 transactions in a single day. 

\subsection{The AST Adversarial Model}
Now we detail the actions and rewards of the adversarial agent (fraudster) of figure \ref{ASTSystem}. Table \ref{AmtStatsTable} shows merchant categories available in our dataset and the associated statistics on overall and fraud transaction amounts. 
\begin{table}[h!]
{\footnotesize
	\begin{center}
		\begin{tabular}
			{|l||c|c|c||c|c|c||}
			\hline
			\multicolumn{1}{|c||}{} & \multicolumn{3}{c||}{Overall} & \multicolumn{3}{c||}{Fraud} \\ \hline
			Merchant Category & Portion & \$ Mean & \$ Stdev & Portion & \$ Mean & \$ Stdev \\
			\hline \hline
			entertainment   & 7\%  & 64  & 64  & 3\%  & 510 & 74 \\ \hline
			food / dining     & 7\%  & 51  & 48  & 3\%  & 122 & 14 \\ \hline
            gas / transport   & 10\% & 64  & 16  & 7\%  & 12  & 5 \\ \hline
            grocery online    & 3\%  & 54  & 23  & 2\%  & 12  & 3 \\ \hline
            grocery in person     & 9\%  & 116 & 52  & 23\% & 313 & 27 \\ \hline
            health / fitness  & 7\%  & 54  & 48  & 2\%  & 20  & 2 \\ \hline
            home            & 9\%  & 58  & 48  & 3\%  & 258 & 47 \\ \hline
            kids / pets       & 9\%  & 58  & 49  & 3\%  & 20  & 3 \\ \hline
            misc online        & 5\%  & 79  & 164 & 13\% & 804 & 87 \\ \hline
            misc in person     & 6\%  & 62  & 134 & 3\%  & 193 & 316 \\ \hline
            personal care   & 7\%  & 48  &  49 & 3\%  & 26  & 12 \\ \hline
            shopping online    & 8\%  & 83  & 237 & 24\% & 994 & 95 \\ \hline
            shopping in person    & 9\%  & 77  & 232 & 10\% & 887 & 131 \\ \hline
            travel          & 3\%  & 112 & 596 & 0\%  & 9   & 2 \\ \hline
		\end{tabular}
    \caption{Comparison of overall vs fraud payment transaction statistics.}
    \label{AmtStatsTable}
	\end{center}}
\end{table}

The table shows that fraudsters prefer merchants such as online shopping. This is  due to their convenience and the ready availability of high value fencible goods.  Fraudsters generally also prefer to accumulate as much fraudulent activity in as short a time period as possible as they know detection is just a matter of time. Fraudsters also generally know the approximate age of a stolen card and therefore can choose the age of a card used in a fraud attempt. These behaviors motivate the following choice of {\bf actions}:
\begin{center}
[card age]$\times$[merchant category]$\times$[amount]$\times$[interval between transaction].
\end{center}

The form of the {\bf reward function} distinguishes AST from other reinforcement learning approaches. It is designed to find failure events as the primary objective, and maximize the path likelihood as a secondary objective \cite{Lee}. We define the following:
\begin{itemize}
    \item {\bf Terminal state}: if fraud is detected or if 10 transactions are completed in a single day.
    
    \item $E$ the {\bf event space} of interest: the fraudster is \underline{not} caught. 
    
    \item $R_E$ the {\bf reward if the fraudster is successful}: given by the total accumulated fraud amount: $ R_E = \sum_{\mbox{fraud trx}} v_k$ where $v_k$ is the amount of fraud transaction $k$.
    
    \item $d$ the {\bf miss distance}: this represents a penalty for being caught. Through experimentation we find that setting this to a large negative constant works best. 
    https://www.overleaf.com/project/60b6f393891eed7268032b1f
    \item $p(a | s)$ the {\bf likelihood of an action in a given state}: this is modeled from the data set by assuming independence of the action components and examining the histograms of each in the dataset where we find:
    {\small
    \begin{align}
        p(n_{tot} = \mbox{total card trx}) & \sim \mbox{exp}(\lambda=1/(390 \mbox { trx})) \nonumber \\
        p(m = \mbox{merch categ}) & \sim \mbox{as in column "Fraud - Portion" of table \ref{AmtStatsTable}} & \nonumber \\
        p(v = \mbox{amount} | m) & \sim \mathcal{N}(\mu_i, \sigma^2_i) \quad \mbox{per table \ref{AmtStatsTable}}, & i \in \mbox{\texttt{enum}(merch categ)} \nonumber \\ 
        p(\delta_i = \mbox{interval} | m) & \sim \mathcal{N}(\mu'_i, \sigma'^2_i) \quad \mbox{similar to table \ref{AmtStatsTable} but not shown}, & i \in \mbox{\texttt{enum}(merch categ)} \nonumber \\ 
        \Rightarrow p(a | s) & = p(n_{tot}) \cdot p(m) \cdot p(v_i | m) \cdot p(\delta_i | m) &
    \label{psgivenaEq}
    \end{align}}
\end{itemize}
With these the reward function is given by
\begin{align}
    r(s, a) & =  \left\{ \begin{array}{ll}
                    R_E & \mbox{if } s \mbox{ is terminal and } s \in E \\
                    -d &  \mbox{if } s \mbox{ is terminal and } s \notin E \\
                    \log(p(a | s)) & \mbox{otherwise}
                        \end{array} \right.
\label{rewardEq}
\end{align}

To reflect the strong preference of fraudsters for quick profit we employ a discount factor that strongly discounts fraud that takes more than a day to collect. Thus we assume a daily $\gamma = 0.2$.

\section{Experiments}
\subsection{Fraud Classifier}
The dataset comprises approximately 1.2 million training transactions across several thousand accounts, and an approximately 500,000 transaction test set, both with fraud rates under 0.4\%. This represents a highly imbalanced dataset. We utilize the synthetic minority oversampling technique (SMOTE) to mitigate this \cite{Chawla}, \cite{Pozzolo}. We first oversample the fraud transactions in our dataset to where they form 10\% of the samples, then create additional synthetic data such that our final training set is 1/3 fraud and 2/3 non-fraud. This approach helps to improve the classifier’s predictive capabilities \cite{Brownlee}.

For simplicity and transparency, we chose to use a logistic regression model with the limited features detailed in section \ref{systemSection} as our classifier. With this classifier we get the following confusion matrix on our validation dataset:
\begin{itemize}
    \item Accuracy: 96.8\% -- these are transactions correctly identified as either fraud or not fraud.
    \item Decline rate: 3.85\% -- these are good transactions mis-identified as potential fraud and erroneously declined and represent missed revenue for merchants.
    \item Uncaught fraud rate: 0.1\% -- these are fraudulent transactions not caught by the classifier. Of the attempted fraud this represents about 26\% (meaning the classifier catches about 3/4 of attempted fraud). 
\end{itemize}
Note that the above values are in line with commercial fraud detection performance. 

\subsection{AST Experiments}
We employ algorithm (\ref{alg:AST}) to estimate the Q function associated with this reinforcement learning environment.  
\begin{algorithm}[t]
\caption{AST $Q$-learning}
\centering
\label{alg:AST}
\begin{algorithmic}[1]
    \Require Action space $\mathcal{A}=$[card age] $\times$ [merch. categ.] $\times$ [amount]$\times$[interval bet transactions], action selection rule $\pi(\cdot; \cdot) \sim U[\mathcal{A}]$, step size $\alpha \in (0,1]$, max daily transactions $T = 10$
    \State Initialize $Q(s,a)$ for all $s \in \mathcal{S}, a \in \mathcal{A}$
    \For{each episode, $n = 0, \ldots, N$}
        \State Initialize the state, $s_0$: agent randomly selects a new or old payment card account to use, daily transactions are set to 0, and the fraud indicator set to 0.
        \For{$t = 0, \ldots, T$}
            \State $a_t \gets \pi(s_t; Q)$
            \State Determine $\log p(s_t | a)$ according to equation (\ref{psgivenaEq})
            \State Classify the resultant payment transaction as fraud / no fraud
            \State Determine reward $r_t$ according to equation (\ref{rewardEq})
            \State Observe next state $x_{t+1}$
            \State $Q(x_t,a_t) \gets Q(x_t,a_t) + \alpha (r_t + \gamma \min_{a' \in \mathcal{A}} Q(x_{t+1},a') - Q(x_t,a_t))$
        \EndFor
    \EndFor
\end{algorithmic}
\end{algorithm}

\begin{figure*}[ht]
    \centering
    \includegraphics[scale=0.25]{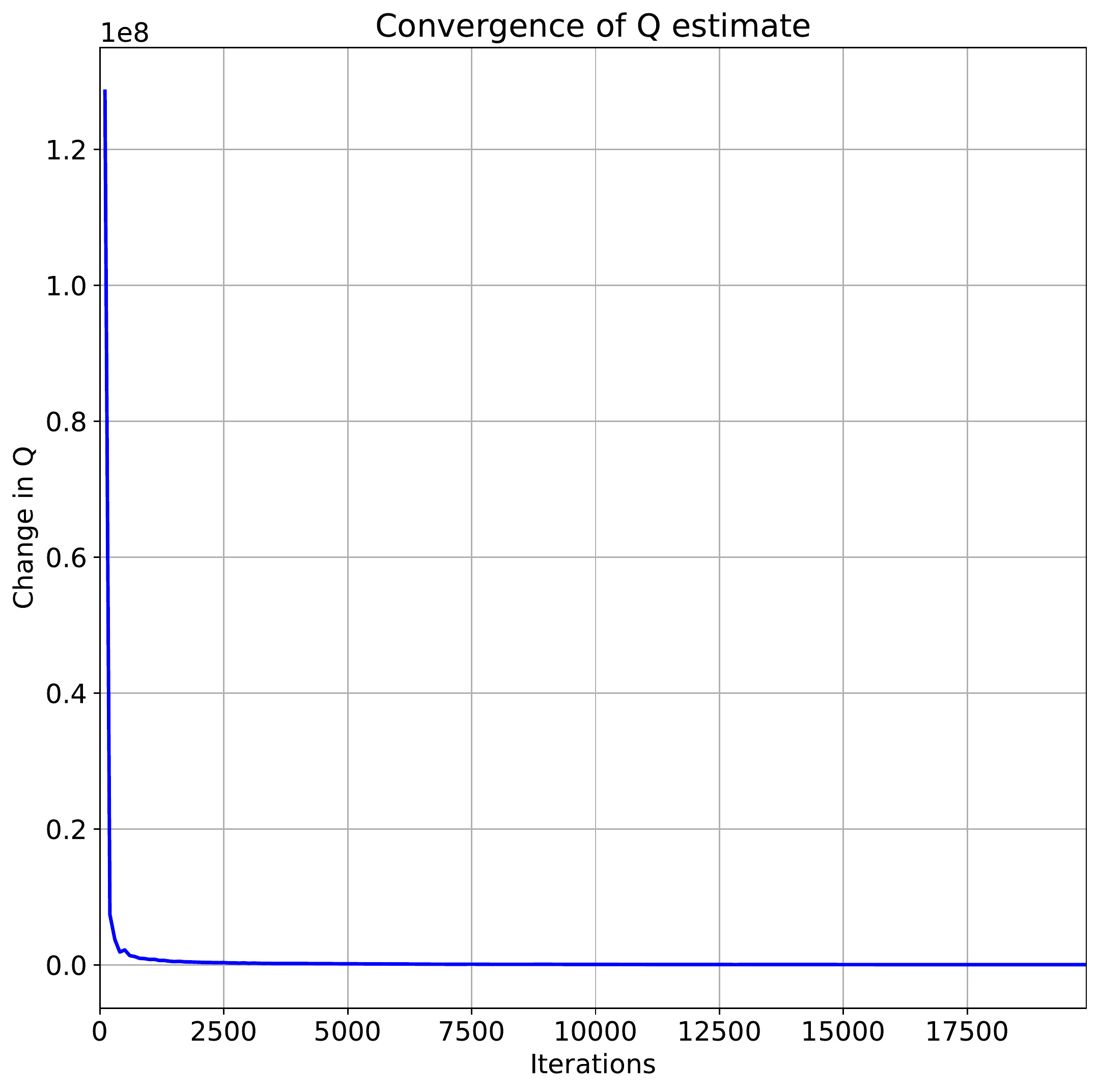}
    \caption{Convergence of the $Q$-function for experiment 1.}
    \label{Qconvergence2}
\end{figure*}

For our first experiment we discretize the action space as
\begin{itemize}
    \item Initial card total transactions: \{10 (new card), 500 (slightly over average age card)\}.
    \item Merchant categories: \{entertainment, shopping online, shopping in person\}. These are chosen due to their affinity to fraud activity (fencible products, potential for high \$ ticket items)
    \item Amounts: \{\$10, \$100, \$1,000\}
    \item Intervals between transactions: \{1 min, 50 min, 150 min\} (all same day transactions)
\end{itemize}
This represents an action space of size $|\mathcal{A}| = 54$. Figure \ref{Qconvergence2} shows the convergence of the $Q$ function as measured by the Frobenius norm of the change over iterations. The resultant most likely fraud path from the learned policy is as shown in Table \ref{policy1}.

\begin{table}[h!]
{\footnotesize
	\begin{center}
		\begin{tabular}
			{|r||l|r|r||l|r|r||}
			\hline
			\multicolumn{1}{|c||}{} & \multicolumn{3}{c||}{New Card} & \multicolumn{3}{c||}{Middle Age Card} \\ \hline
			Daily tran & Category & \$ Amount & Int. (min) & Category & \$ Amount & Int. (min)  \\
			\hline \hline
			1 & Entertainment & 100 & 150 & Shopping online & 100 & 50 \\ \hline
			2 & Shopping online & 100 & 150 & Shopping online & 100 & 150 \\ \hline
			3 & Shopping online & 100 & 1 & Shopping online & 100 & 50  \\ \hline
			4 & Shopping online & 100 & 1 & Shopping online & 100 & 150 \\ \hline
			5 & Shopping online & 100 & 1 & Shopping online & 100 & 1 \\ \hline
			6 & Shopping online & 100 & 1 & Shopping online & 100 & 50 \\ \hline
			7 & Shopping in person & 100 & 50 & Shopping online & 100 & 1 \\ \hline
			8 & Shopping in person & 100 & 1 & Shopping online & 100 & 50 \\ \hline
			9 & Shopping in person & 100 & 50 & Shopping online & 100 & 150 \\ \hline
			10 & Entertainment & 100 & 150 & Entertainment & 100  & 150 \\ \hline
		\end{tabular}
    \caption{AST most likely fraud path detected for the first experiment (no new business rules).}
    \label{policy1}
	\end{center} }
\end{table}
Inspection of the results reveals that the fraudster will always revert to a moderate size transaction amount (\$100) and mostly use the online shopping channel. There is a diverse range of intervals utilized. We might intuit from this that the fraud classifier does a good job of detecting "typical" fraud such as large amount transactions. But if a fraudster chose to pursue a repetitive moderate amount transaction strategy the system may not catch it. 

It is difficult to modify or shape the logistic regression classifier behavior to attempt to mitigate this fraud strategy. But the business rules are well equipped to handle this. A rule can be added to monitor for long successions of similar transaction sizes. In fact we do this and re-run the experiment and obtain the most likely path strategy of Table \ref{policy2}. Most notably here we see the fraudster must modify the transaction amounts in this case.
\begin{table}[h!]
{\footnotesize
	\begin{center}
		\begin{tabular}
			{|r||l|r|r||l|r|r||}
			\hline
			\multicolumn{1}{|c||}{} & \multicolumn{3}{c||}{New Card} & \multicolumn{3}{c||}{Middle Age Card} \\ \hline
			Daily tran & Category & \$ Amount & Int. (min) & Category & \$ Amount & Int. (min)  \\
			\hline \hline
			1 & Shopping in person & 100 & 150 & Shopping online & 100 & 50 \\ \hline
			2 & Shopping online & 1,000 & 150 & Shopping in person & 100 & 150 \\ \hline
			3 & Shopping online & 1,000 & 1 & Shopping online & 1,000 & 50  \\ \hline
			4 & Entertainment & 1,000 & 1 & Shopping online & 1,000 & 150 \\ \hline
			5 & Shopping online & 100 & 1 & Shopping online & 100 & 1 \\ \hline
			6 & Shopping online & 1,000 & 50 & Shopping online & 1,000 & 50 \\ \hline
			7 & Shopping in person & 100 & 1 & Shopping in person & 100 & 1 \\ \hline
			8 & Entertainment & 1,000 & 1 & Shopping online & 100 & 50 \\ \hline
			9 & Entertainment & 100 & 150 & Entertainment & 100 & 150 \\ \hline
			10 & Shopping in person & 100 & 150 & Entertainment & 1,000  & 150 \\ \hline
		\end{tabular}
    \caption{AST most likely fraud path detected for the second experiment with rule preventing lengthy successions of similar transaction amounts.}
    \vspace{-0.25in}
    \label{policy2}
	\end{center} }
\end{table}

This second experiment also plants the seed for subsequent study, namely utilizing this framework in an interactive game setting where the fraudster develops a strategy, the system mitigates it, and iterations continue until a more refined fraud detection system is developed. 

\vspace{-0.05in}
\section{Conclusions and Future Work}
In this report our contributions include:
\begin{itemize}

    \item Casting the Adaptive Stress Testing paradigm as an adversarial learning problem for payment card fraud detection.
    
    \item Creating a simple but realistic payment card fraud detection system that incorporates a statistical classifier and a set of business rules to mitigate fraud and that resembles systems used in practice by card issuers. 

    \item Showing the most likely fraud path as a sequence of actions that can be taken by a fraudster to maximize their profit and providing an intuitive explanation why it works. 

    \item Modifying the fraud detection system through the addition of a business rule to prevent the identified strategy and showing that it results in a change in fraudster strategy.

    \item Motivating continued study of the use of AST in an interactive game-like setting between the system and fraudster to fine-tune the system's performance.
    
\end{itemize}

Future work would mainly revolve around exploring the interactive game-setting implementation of this adversarial learning approach. Some areas of specific investigation might include: what is a process for automating a business rule in response to a fraud strategy, and how to we protect against unintended consequences of new business rules. Also, the fraud classifier is ripe for additional sophistication utilizing more data fields and features and more powerful models such as a neural network. 


\bibliography{refs}
\bibliographystyle{plain}

\end{document}